\pdfoutput=1

\documentclass[11pt]{article}

\usepackage{acl}

\usepackage{times}
\usepackage{latexsym}

\usepackage[T1]{fontenc}

\usepackage[utf8]{inputenc}

\usepackage{microtype}

\newcommand{\boldX}{{\boldsymbol{X}}}

\newcommand{\boldd}{{\boldsymbol{d}}}

\newcommand{\boldx}{{\boldsymbol{x}}}

\usepackage{bbm}
\usepackage[ruled, algo2e, vlined]{algorithm2e}

\SetCommentSty{mycommfont}

\usepackage[utf8]{inputenc} 
\usepackage[T1]{fontenc}    
\usepackage{hyperref}       
\usepackage{url}            
\usepackage{booktabs}       
\usepackage{amsfonts}       
\usepackage{nicefrac}       
\usepackage{microtype}      
\usepackage{xcolor}         

\hypersetup{colorlinks, citecolor=blue}

\usepackage[ruled,vlined]{algorithm2e}
\usepackage{setspace}

\usepackage{epsfig,bm,subfigure,graphicx,color}
\usepackage{subfigure}
\usepackage{wrapfig}

\usepackage{here}

\usepackage[normalem]{ulem}

\usepackage{amsmath,amssymb,amsfonts,mathtools}
\usepackage{bbm}
\usepackage{lscape}

\usepackage{tcolorbox}

\DeclareMathOperator*{\minimize}{minimize}

\title{Word Tour: One-dimensional Word Embeddings via \\ the Traveling Salesman Problem}

\author{Ryoma Sato \\
  Kyoto University / RIKEN AIP \\
  \texttt{r.sato@ml.ist.i.kyoto-u.ac.jp}}

\begin{document}
\maketitle
\begin{abstract}
Word embeddings are one of the most fundamental technologies used in natural language processing. Existing word embeddings are high-dimensional and consume considerable computational resources. In this study, we propose \textsc{WordTour}, unsupervised \emph{one-dimensional} word embeddings. To achieve the challenging goal, we propose a decomposition of the desiderata of word embeddings into two parts, completeness and soundness, and focus on soundness in this paper. Owing to the single dimensionality, \textsc{WordTour} is extremely efficient and provides a minimal means to handle word embeddings. We experimentally confirmed the effectiveness of the proposed method via user study and document classification.
\end{abstract}

\section{Introduction}

Word embeddings are one of the most thriving techniques in natural language processing and are used in various tasks, including word analogy \cite{mikolov2013linguistic, pennington2014glove}, text classification \citep{kim2014convolutional, kusner2015from, shen2018baseline}, and text similarity \cite{arora2017simple, yokoi2020word}. Existing word embeddings are in high-dimensional spaces. Although high dimensionality offers representational power to word embeddings, it also has the following drawbacks: (1) \textbf{Memory inefficiency.} High-dimensional word embeddings require the storage of many floating-point values, and they consume considerable memory space. For instance, the $300$-dimensional GloVe with $400$k words consumes $1$ GB of memory. This hinders the application of word embeddings in edge devices \citep{raunak2019effective, jurgovsky2016evaluating, joulin2016fasttextzip}. (2) \textbf{Time inefficiency.} The high dimensionality also increases the time consumption owing to many floating-point arithmetic operations. (3) \textbf{Uninterpretability.} It is not straightforward to visualize high-dimensional embeddings. Projections to low dimensional spaces, e.g., by t-SNE and PCA, lose some information, and it is difficult to control and interpret the aspects that these projections preserve. Besides, word embeddings are sparse in high-dimensional space, and for a small perturbation $\varepsilon \in \mathbb{R}^d$, it is not clear what $\boldx_{\text{cat}} + \varepsilon$ represents, e.g., when creating adversarial examples \cite{lei2019discrete} and data augmentation \cite{qu2021coda}.

In this study, we propose \textsc{WordTour}, unsupervised \emph{one dimensional} word embeddings.
In contrast to high-dimensional embeddings, \textsc{WordTour} is \textbf{memory efficient}. It does not require storing even a single floating-point value; instead, it stores only the order of words. \textsc{WordTour} with $40$k words consumes only $300$ KB memory, which is the same space as the space for storing a list of the words. Memory efficiency enables applications in low-resource environments. \textsc{WordTour} is \textbf{time efficient} as well. It can compare words in a single operation whereas traditional embeddings require hundreds of floating-point operations for a single comparison. In addition, it can retrieve similar words by simply looking up the surrounding words in a constant time and can efficiently compare documents using a blurred bag of words, as we will show in the experiments. These features are also advantageous in low resource environments. In addition, \textsc{WordTour} is \textbf{interpretable} owing to its single dimensionality. It is straightforward to visualize the one dimentional embeddings without any information loss. Besides, we can always interpret the perturbed word embedding as we can interpret the perturbed image pixels. 
In brief, \textsc{WordTour} provides a \emph{minimal} means to handle word embeddings.

However, words are inherently high-dimensional, and it is impossible to capture \emph{all} semantics in one dimension. To tackle this challenge, we propose to decompose the desiderata of word embeddings into two components: soundness and completeness. \textsc{WordTour} gives up completeness, focuses on soundness, and thereby realizes meaningful one dimensional embeddings for some, if not all, applications. We formulate the optimization of sound word embeddings as the traveling salesman problem and solve it using a highly efficient solver. In the experiments, we confirm that \textsc{WordTour} provides high-quality embeddings via qualitative comparison, user studies, and document classification.

\begin{tcolorbox}[colframe=gray!20,colback=gray!20,sharp corners]
\textbf{Reproducibility}: Our code and obtained embeddings are available at \url{https://github.com/joisino/wordtour}.
\end{tcolorbox}

\section{Backgrounds}

\subsection{Notations}

Let $\mathcal{V}$ be the set of words in a vocabulary, and $n = |\mathcal{V}|$ be the number of words. Let $[n] = \{1, 2, \cdots, n\}$ and let $\mathcal{P}([n])$ be the set of permutations of $[n]$. 

\subsection{Problem Definition}

We are given off-the-shelf word embeddings $\boldX = [\boldx_1, \cdots, \boldx_n]^\top \in \mathbb{R}^{n \times d}$, such as word2vec and GloVe. We assume that the embeddings completely represent the semantics of the words, but they are high-dimensional, e.g., $d = 300$. We aim to create an ordering of $\mathcal{V}$ such that the order preserves the structure of the given embeddings. The problem is defined as follows:

\begin{tcolorbox}[colframe=gray!20,colback=gray!20,sharp corners]
\textbf{Problem Definition.}

\textbf{Given:} Word embedddings $\boldX \in \mathbb{R}^{n \times d}$.

\textbf{Output:} Word ordering $\sigma^* \in \mathcal{P}([n])$.
\end{tcolorbox}

In full generality, it may be possible to model the real-value positions. However, in this paper, we solely consider the order of the words. That is, the words are equally spaced in the one-dimensional space. This formulation makes the embedding simpler and lighter, while still being sufficiently powerful.

\section{Word Tour}

\begin{figure}[t]
\centering
\includegraphics[width=0.8\hsize]{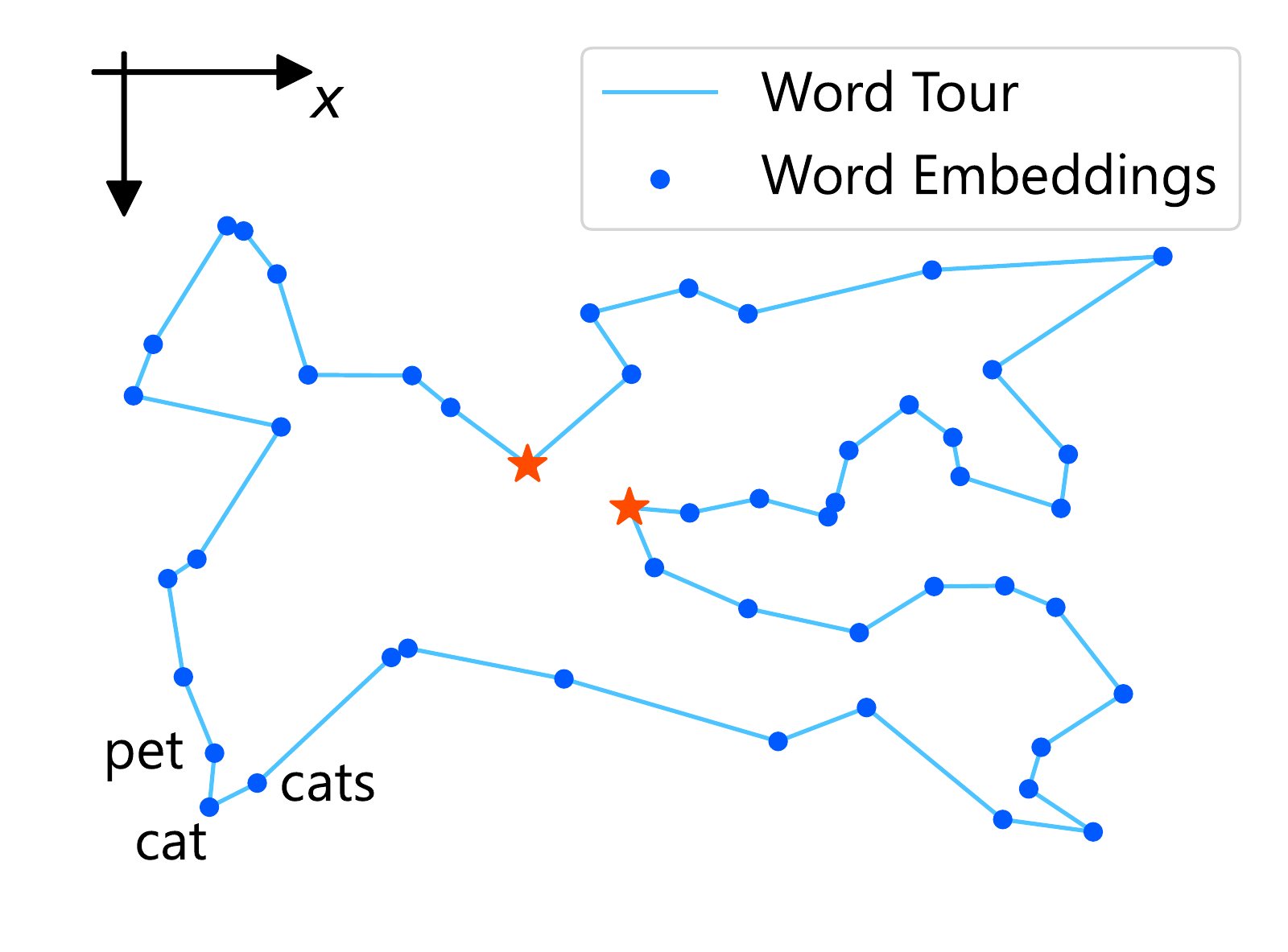}
\vspace{-0.15in}
\caption{\textbf{Illustration of \textsc{WordTour}}. Each dot represents a word with its coordinates as the embedding vector.}
\label{fig: illust}
\vspace{-0.2in}
\end{figure}

In this section, we introduce our proposed method, \textsc{WordTour}. Ideally, we would like to preserve all the semantics in our one-dimensional embeddings. However, such ideal embeddings are unlikely to exist because the relations between words are inherently high-dimensional. Indeed, although existing studies have attempted to reduce the dimensionality of word embeddings, they require at least tens of dimensions \citep{raunak2019effective, acharya2019online} and several dimensions even in non-Euclidean spaces \citep{nickel2017poincare, tifrea2019poincare}. These results indicate that ideal 1D embeddings do not exist. Therefore, we make a compromise. We decompose the desiderata of word embeddings into the following two categories: 
\begin{description}
    \item[Soundness] Close embeddings should have semantically similar meanings.
    \item[Completeness] Semantically similar words should be embedded closely.
\end{description}
In \textsc{WordTour}, we give up the latter condition and focus on the former condition. For instance, the two red stars in Figure \ref{fig: illust} are distant in the order, although they are semantically similar. \textsc{WordTour} accepts such inconsistency. Owing to the incompleteness, \textsc{WordTour} may fail some applications of word embeddings, such as word analogy and relation extraction. Nevertheless, \textsc{WordTour} still has some other applications, such as word replacement and document retrieval. Indeed, \textsc{WordTour} may overlook some relevant documents because they may embed relevant words far apart. However, the close documents found by \textsc{WordTour} are indeed close owing to soundness. These insights indicate that there exist one-dimensional embeddings that are useful for some, if not all, applications.

\begin{table*}[t]
    \centering
    \caption{\textbf{Examples of segments}. Each row represents a segment. (a--d) Segments around ``cat.'' (e--h) Segments around ``concept.'' (i--o) Random segments of \textsc{WordTour}. \textsc{WordTour} provides smooth orderings.}
    \vspace{-0.1in}
    \scalebox{0.55}{
\begin{tabular}{lc|ccccccccccc} \toprule
 & Methods & \multicolumn{11}{c}{Segments} \\ \midrule
(a) & \textsc{WordTour} & sniff & sniffing & sniffer & dogs & dog & cat & cats & pets & pet & stray & errant \\
(b) & RandProj & loire & sayings & nn & trooper & referendum & cat & exceeded & traces & freestyle & mirrored & bloomberg \\
(c) & PCA1 & mm & asylum & kohl & presents & expressed & cat & sichuan & denmark & counted & corporations & hewitt \\
(d) & PCA4 & 1.46 & puzzles & 940 & coexist & locations & cat & att & winners & perth & colgate & sohail \\ \midrule
(e) & \textsc{WordTour} & assumption & assumptions & notions & notion & idea & concept & concepts & ideas & thoughts & feelings & emotions \\
(f) & RandProj & entertaining & 42,000 & kursk & embarrassment & ingrained & concept & berezovsky & cg & guillen & excerpts & roofs \\
(g) & PCA1 & neighboring & branches & argued & manhattan & 1998 & concept & share & pending & response & airlines & fort \\
(h) & PCA4 & 2:00 & hksar & hashim & provider & straining & concept & inducing & fightback & unsettled & bavaria & sign \\ \midrule
(i) & \textsc{WordTour} & wireless & broadband & 3g & cdma & gsm & handset & handsets & smartphones & smartphone & blackberry & tablet \\
(j) & \textsc{WordTour} & gun & weapon & weapons & arms & arm & leg & legs & limbs & limb & prosthetic & make-up \\
(k) & \textsc{WordTour} & federalist & libertarian & progressive & liberal & conservative & conservatives & liberals & democrats & republicans & gop & republican \\
(l) & \textsc{WordTour} & cordial & amicable & agreeable & mutually & beneficial & detrimental & harmful & destructive & disruptive & behaviour & behavior \\
(m) & \textsc{WordTour} & 15th & 14th & 13th & 12th & 10th & 11th & 9th & 8th & 7th & 6th & 5th \\
(n) & \textsc{WordTour} & suspicions & doubts & doubt & doubted & doubting & doubters & skeptics & skeptic & believer & believers & adherents \\
(o) & \textsc{WordTour} & molten & magma & lava & basalt & sandstone & limestone & granite & marble & slab & slabs & prefabricated \\\bottomrule
    \end{tabular}
    }
    \label{tab: example}
    \vspace{-0.1in}
\end{table*}

A natural criterion for soundness is that consecutive words in the ordering should be close to one another in the original embedding space. We formulate the problem as follows:
\begin{align} \label{eq: tsp}
\minimize_{\sigma \in \mathcal{P}([n])} \|\boldx_{\sigma_1} - \boldx_{\sigma_n}\| + \sum_{i = 1}^{n-1} \|\boldx_{\sigma_i} - \boldx_{\sigma_{i+1}}\|.
\end{align}
We treat the ordering as a cycle, not a path, by adding term $\|\boldx_{\sigma_1} - \boldx_{\sigma_n}\|$. The rationale behind this design is that we would like to treat all words symmetrically and would like the boundary words to have the same number of neighbors as the non-boundary words. 

In formulation \eqref{eq: tsp}, we adopt the $L_2$ norm for simplicity. However, our formulation is agnostic to the distance function. When a corpus is at hand, we can also use the number of co-occurrences, i.e., $\sum_{i} \#\text{co-occurrences of } (\sigma_i, \sigma_{i+1})$ , as the cost function. We leave investigating other modelings as future work and focus on the $L_2$ cost in this paper.

The optimization problem \eqref{eq: tsp} is an instance of the traveling salesman problem (TSP), which is NP-hard. 
As the problem size is relatively large in our case, for instance, $n = 40\,000$, it may seem impossible to solve the problem. However, in practice, highly efficient TSP solvers have been developed. Among others, we employ the LKH solver \citep{lkh_webpage}, which implements the Lin Kernighan algorithm \citep{lin1973effective, helsgaun2000effective} in a highly efficient and effective manner. The LKH solver performs a restricted local search based on a guide graph constructed using the dual problem. \citet{lkh_webpage} reported that the LKH solver \emph{exactly} solved an instance with as many as $109\,399$ cities. In addition, several effective algorithms for computing lower bounds provide theoretical guarantees for the quality of a solution. We employ the one-tree lower bound \citep{helsgaun2000effective} implemented in the LKH solver to compute the lower bounds of the optimum value. As a tour is a special case of a one-tree, the minimum cost one-tree is a provable lower bound of the TSP problem. The algorithm searches for a potential vector for a tight lower bound by gradient ascent. \textsc{WordTour} computes a near-optimal solution of Problem \eqref{eq: tsp} by the LKH solver and uses the solution as the word order, i.e., the word embeddings.

\section{Experiments}

We experimentally validated the effectiveness of \textsc{WordTour}. We used a Linux server with Intel Xeon E7-4830 v4 CPUs in the experiments.

\subsection{Computing Embeddings}

We used $300$-dimensional GloVe embedding with the first $40\,000$ words as the input embeddings $\{\boldx_v\}$. The objective value of the solution obtained by LKH was $236882.314$, and the lower bound proved by LKH was $236300.947$. Therefore, the cost of the obtained tour is guaranteed to be at most $1.003$ of the optimum.
The resulting embedding file is $312$ KB, which is sufficiently light to be deployed in low-resource environments.

\subsection{Qualitative Comparison}

We use the following baselines: (1) \textbf{RandProj} randomly samples a direction $\boldd \in \mathbb{R}^d$ and orders the embeddings in ascending order of $\boldd^\top \boldx_i$. This method extracts a specific aspect $\boldd$ of the input embeddings. (2) \textbf{PCA-1} orders in ascending order of the top PCA component. (3) \citet{mu2018all} reported that a few leading PCA components were not informative. Therefore, \textbf{PCA-4} orders words by the fourth PCA component.

As we cannot show the entire tour owing to space constraints, we sample and list some random segments in Table \ref{tab: example}. It is observed that \textsc{WordTour} provides the most natural ordering, and the consecutive words are semantically similar in \textsc{WordTour}. Notably, \textsc{WordTour} almost recovers the order of ordinals without explicit supervision (Table \ref{tab: example} (m)).

\subsection{Assesment via Crowdsourcing}

\begin{figure}[t]
\centering
\includegraphics[width=0.8\hsize]{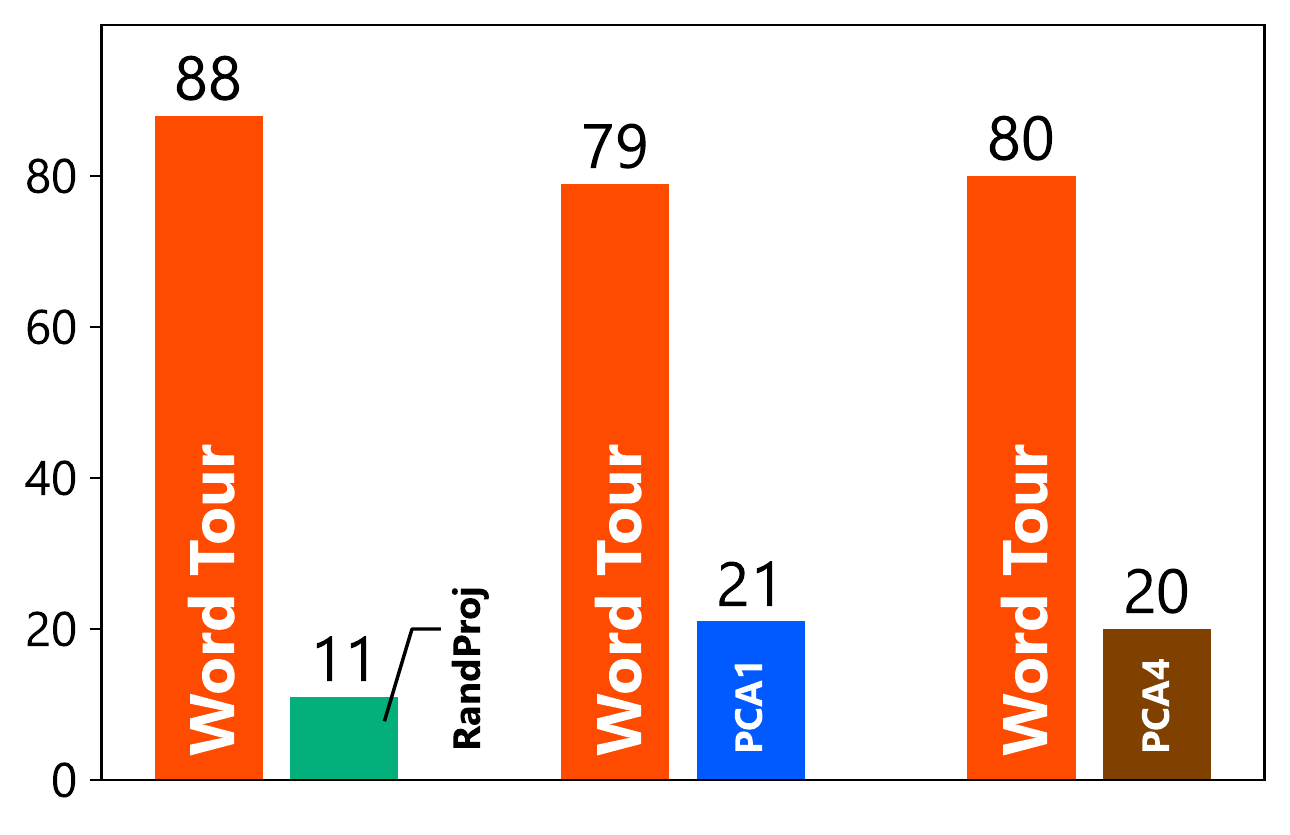}
\vspace{-0.1in}
\caption{\textbf{Results of the user study}. Each bar represents the number of times each method was selected within $100$ trials. One trial was not completed in \textsc{WordTour} vs. RandProj, which led to $99$ trials in the first comparison.}
\label{fig: crowd}
\vspace{-0.2in}
\end{figure}

We conducted a user study at Amazon Mechanical Turk to confirm the effectiveness of \textsc{WordTour}. Specifically, to compare two word ordering $\sigma, \tau \in \mathcal{P}([n])$, we randomly sample a reference word $v \in \mathcal{V}$, retrieve the next words of $v$ in $\sigma$ and $\tau$, and ask a crowdworker which word is more similar to the reference word $v$. We repeated this process $100$ times for each pair of embeddings. Figure \ref{fig: crowd} shows the number of times each embedding was selected. This clearly shows that \textsc{WordTour} aligns with human judgment.

\subsection{Document Retrieval}

\begin{figure}[t]
\centering
\includegraphics[width=\hsize]{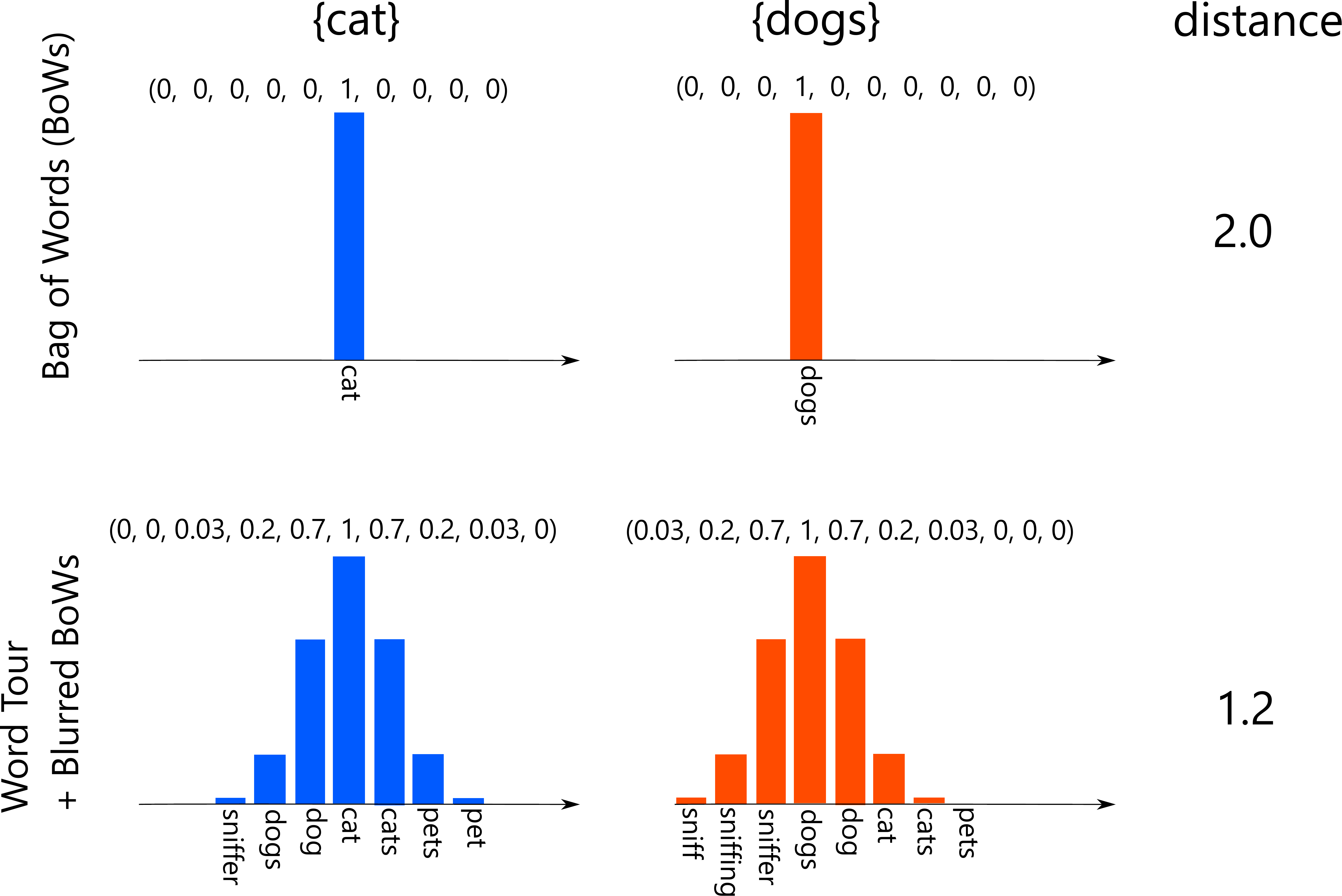}
\vspace{-0.1in}
\caption{Document comparison by \textsc{WordTour}. This figure illustrates the case in which a document is composed of a single word. When more than one word is in a document, the blurred BoW will be multimodal.}
\label{fig: compare}
\vspace{-0.1in}
\end{figure}

In this section, we evaluate the effectiveness of word embeddings in document classification. The most straightforward approach to compare two documents is the bag of words (BoW), which counts common and uncommon words in documents. However, this approach cannot capture the similarities of the words. In 1D embeddings, neighboring words are similar, although they are not exactly matched in BoW. To utilize this knowledge, we use blurred BoW, as shown in Figure \ref{fig: compare}. Specifically, we put some mass around the words in a document to construct the blurred BoW vector. We employ a Gaussian kernel for the mass amount and use \textsc{WordTour}, RandProj, PCA1, and PCA4 for the orderings. We normalize the BoW and blurred BoW vectors with the $L_1$ norm and compute the distance between two documents using the $L_1$ distance of the vectors. The blurred BoW can be computed in $O(wn)$ time, where $n$ denotes the number of words in a document and $w$ is the width of the filter. We used $w = 10$ in the experiments. We also use \textbf{word mover's distance (WMD)} \cite{kusner2015from} as a baseline, which is one of the most popular word-embedding-based distances. We used $300$-dimensional GloVe for WMD. WMD requires $O(n^3 + n^2 d)$ computation because of the optimal transport formulation, where $n$ denotes the number of words in a document and $d$ is the number of dimensions of word embeddings. The performance of WMD can be seen as an expensive upper bound of BoW and blurred BoW. We used five datasets: ohsumed \cite{joachims1998text}, reuter \cite{sebastiani2002machine}, 20news \cite{lang1995news}, Amazon \cite{blitzer2007biographies}, and classic \cite{classic}. We remove the duplicated documents following \cite{sato2021reevaluating}. The details of the datasets are provided in the Appendix. We evaluated the performance using the $k$-nearest neighbor error. We used the standard test dataset if it existed (for instance, based on timestamps) and used five random train/test splits for the other datasets\footnote{The seeds are fixed and reported in the GitHub repository.}. We report the standard deviations for five-fold datasets.

\begin{table}[t]
    \centering
    \caption{\textbf{Document classification errors}. \emph{Lower is better}. The time row reports the average time to compare the two documents. \textsc{WordTour} performs the best in the blurred BoW family.}
    \vspace{-0.1in}
    \scalebox{0.55}{
\begin{tabular}{lccccc} \toprule
 & ohsumed & reuter & 20news & amazon & classic \\ \midrule
BoW & 48.1 & 5.6 & 35.4 & 11.4 $\pm$ 0.4 & 5.1 $\pm$ 0.3 \\
Time & 39 ns & 23 ns & 35 ns & 21 ns & 23 ns \\ \midrule
\textsc{WordTour} & \textbf{47.2} & \textbf{4.6} & \textbf{34.1} & \textbf{10.1 $\pm$ 0.3} & \textbf{4.6 $\pm$ 0.1} \\
RandProj & 47.9 & 5.4 & 35.4 & 11.3 $\pm$ 0.3 & 5.1 $\pm$ 0.3 \\
PCA1 & 47.8 & 5.7 & 35.5 & 11.4 $\pm$ 0.6 & 5.1 $\pm$ 0.3 \\
PCA4 & 48.1 & 5.6 & 35.4 & 11.6 $\pm$ 0.5 & 5.1 $\pm$ 0.4 \\
Time & 206 ns & 142 ns & 312 ns & 185 ns & 150 ns \\ \midrule
WMD & 47.5 & 4.5 & 30.7 & 7.6 $\pm$ 0.3 & 4.2 $\pm$ 0.3 \\
Time & 3.5 $\times 10^6$ ns & 2.2 $\times 10^6$ ns & 5.1 $\times 10^6$ ns& 1.2 $\times 10^7$ ns & 1.9 $\times 10^6$ ns \\ \bottomrule
    \end{tabular}
    }
    \label{tab: classification}
    \vspace{-0.15in}
\end{table}

The results are shown in Table \ref{tab: classification}. Although \textsc{WordTour} is less effective than WMD, it is much faster than WMD and more effective than other 1D embeddings. Recall that the 1D embeddings are designed for low-resource environments, where WMD may be infeasible. \textsc{WordTour} offers an efficient approach while integrating the similarities of the words.

\section{Related Work}

\citet{raunak2019effective} and \citet{jurgovsky2016evaluating} proposed a postprocessing method to reduce the number of dimensions of the off-the-shelf word embeddings. However, existing methods require at least five to tens of dimensions. To the best of our knowledge, this study is the first to obtain large-scale 1D word embeddings. \citet{nickel2017poincare} proposed to embed words into hyperbolic spaces and drastically reduce the number of required dimensions. FastText.zip \citep{joulin2016fasttextzip} quantizes and prunes word embeddings for memory-efficient text classification. Although FastText.zip saves considerable memory consumption without harming downstream tasks, it prunes words that are irrelevant to text classification, whereas we aim to retain the original vocabulary in this work. \citet{ling2016word} and \citet{tissier2019near} proposed to quantize general word embeddings. Although they save considerable memory and time complexity with no considerable performance degradation, they still consume a few orders of magnitude more memory than 1D embeddings, and they are sparse in the embedding space and require more time than \textsc{WordTour} to compare documents and search similar words.

\section{Conclusion}

In this study, we proposed \textsc{WordTour}, a 1D word embedding method. To realize 1D embedding, we decompose the requirement of word embeddings into two parts and impose only one constraint in which the consecutive words should be semantically similar. We formulate this problem using the TSP and solve it with a state-of-the-art solver. Although the TSP is NP-hard, the effective solver solves the optimization almost optimally and provides effective 1D embeddings. We confirmed its effectiveness via crowdsourcing and document classification.

\section*{Acknowledgements}
This work was supported by JSPS KAKENHI GrantNumber 21J22490.

\bibliography{citation}
\bibliographystyle{acl_natbib}

\appendix

\begin{table*}[t]
    \centering
    \caption{Dataset statistics.}
    \begin{tabular}{ccccccccc} \toprule
         & ohsumed & reuter & 20news & amazon & classic \\ \midrule
        Number of documents & 7497 & 7585 & 18776 & 7854 & 6778 \\
        Number of training documents & 3268 & 5413 & 11265 & 5497 & 4744 \\
        Number of test documents & 4229 & 2172 & 7511 & 2357 & 2034 \\
        Size of the vocabulary & 12144 & 13761 & 28825 & 21816 & 12904 \\
        Unique words in a document & 94.5 & 63.0 & 137.1 & 201.8 & 60.8 \\
        Number of classes & 10 & 8 & 20 & 4 & 4 \\ 
        Split type & one-fold & one-fold & one-fold & five-fold & five-fold \\ \bottomrule
    \end{tabular}
    \label{tab: statistics}
\end{table*}

\newpage

\section{Datasets}

Table \ref{tab: statistics} summarizes the statistics of the datasets after preprocessing. Ohsumed \cite{joachims1998text} consists of medical abstracts. Reuter \cite{sebastiani2002machine} and 20news \cite{lang1995news} are news datasets. Amazon \cite{blitzer2007biographies} consists of reviews in amazon.com. Classic \cite{classic} consists of academic papers. The datasets are retrieved from \url{https://github.com/mkusner/wmd}.

\section{Usage of LKH}

We used LKH version 3.0.6, with parameter $\text{PATCHING\_C} = 3, \text{PATCHING\_A} = 2$, which are the default parameters.
As the LKH solver accepts only integral values, we multiply the actual distance by $10^3$ and round down the values before we feed them into the LKH solver. The difference caused by this rounding process is negligibly small.

\section{Hyperparameters}

The number $k$ of neighbors in the kNN classification is selected from $\{1, 2, \cdots, 19\}$. The variance of the Gaussian filter in a blurred bag of words is selected from $\{0.01, 0.1, \cdots, 1000\}$. We selected the hyperparameters using a $5$-fold cross-validation and retrained the kNN model using the chosen hyperparameters and entire training dataset.

\end{document}